\documentclass{article}

\usepackage[dblblindworkshop, final]{neurips_2025}
\workshoptitle{AI for Tabular Data}

\usepackage[utf8]{inputenc} % allow utf-8 input
\usepackage[T1]{fontenc}    % use 8-bit T1 fonts
\usepackage{hyperref}       % hyperlinks
\usepackage{url}            % simple URL typesetting
\usepackage{booktabs}       % professional-quality tables
\usepackage{amsfonts}       % blackboard math symbols
\usepackage{nicefrac}       % compact symbols for 1/2, etc.
\usepackage{microtype}      % microtypography
\usepackage{xcolor}         % colors

\usepackage{amsmath}
\usepackage{amssymb}
\usepackage[disable]{todonotes}
\usepackage{subcaption}

\newcommand{\TabICL}{\textit{TabICL}}
\newcommand{\TabPFNone}{\textit{TabPFN(v1)}}
\newcommand{\TabPFNtwo}{\textit{TabPFN(v2)}}
\newcommand{\TabPFN}{\textit{TabPFN}}

\title{Towards Understanding Layer Contributions in Tabular In-Context Learning Models}

\author{%
  Amir Rezaei Balef$^{1,2,3}$, Mykhailo Koshil$^{2,3}$ and Katharina Eggensperger$^{2,3}$ \\
  $^1$University of Tübingen $^2$ TU Dortmund University \\
  $^3$Lamarr Institute for Machine Learning and Artificial Intelligence \\
 \texttt{\{amir.balef, mykhailo.koshil, katharina.eggensperger\}@tu-dortmund.de} \\
}

\begin{document}

\maketitle

\vspace{-0.2cm}
\begin{abstract}
Despite the architectural similarities between tabular in-context learning (ICL) models and large language models (LLMs), little is known about how individual layers contribute to tabular prediction. In this paper, we investigate how the latent spaces evolve across layers in tabular ICL models, identify potential redundant layers, and compare these dynamics with those observed in LLMs. We analyze \textit{TabPFN} and \textit{TabICL} through the “layers as painters” perspective, finding that only subsets of layers share a common representational language, suggesting structural redundancy and offering opportunities for model compression and improved interpretability.
\end{abstract}

\section{Introduction}
\label{sec:intro}
Tabular in-context learning (ICL) models have demonstrated that transformer-based architectures can achieve state-of-the-art performance on small and medium-sized predictive tabular tasks \citep{erickson2025tabarena}. However, their internal dynamics remain underexplored, and insights from architecturally similar LLM interpretability studies \citep{gromov2025the,sun2025layesaspainters} do not directly transfer due to the differences in inference and training.
Unlike LLMs, prominent ICL models (\TabICL{}, \TabPFN{}) for tabular data do not perform inference in an autoregressive fashion, use attention within the token, and typically do not use positional encodings. Gaining a deeper understanding of how these tabular ICL models function can help identify their strengths, expose their failure modes, and guide future architectural improvements.
Motivated by this goal, we aim to open research in this direction by asking the following question(s):
\textbf{How does data representation evolve in tabular ICL models, and do such models utilize all layers to perform effectively?}  

Here we provide a first step to investigate this question based on three popular tabular ICL models,  \TabPFNone{}~\citep{hollmann-iclr23a}, \TabPFNtwo{}~\citep{hollmann-nature25a}, and \TabICL{}~\citep{qu2025tabicl}. Inspired by the ``Layers as Painters'' framework of \citet{sun2025layesaspainters}, we modify the internal embedding flow within a model to analyze how information is transformed across layers.

\section{Background and Motivation}
\label{sec:background}

We summarize key characteristics of tabular ICL models and findings on layer-wise interpretability.

\textbf{Tabular ICL models} are a particular branch of transformers trained to solve supervised learning problems via in-context learning. In these settings, the input to the model is a support set (train data; feature and target values) and a query (test data; only feature values). The model then predicts the target value for the query, without performing weight updates. Unlike LLMs, tabular ICL models are mostly trained to learn a map between a query token and a label (or value in case of regression).

For the ICL to reach its full potential in transformer models, a large amount of data is required \citep{brown-neurips20}. To satisfy this need, most ICL models 

are directly trained to approximate Bayesian inference on synthetically generated predictive tasks.
\TabPFNone{} \citep{hollmann-iclr23a} was the first model operating in such fashion on tabular data, where the backbone is a vanilla transformer. \TabPFNtwo{} \citep{hollmann-nature25a} improves by adding an attention mechanism within the tokens in addition to cross-tokens attention. \TabICL{} \citep{qu2025tabicl} shares a similar backbone with \TabPFNtwo{}, but additionally introduces a transformer-based compression that efficiently transforms rows into semantically rich embeddings. Specifically, \TabICL{} employs a two-stage architecture. The first stage, ``tabular embedding'', encodes table rows into dense vector representations while explicitly accounting for the inherent column–row structure through column-wise and row-wise interactions. The second stage, ``ICL prediction'', uses these embeddings along with their corresponding labels to make predictions.

\textbf{LLM's hidden state dynamics} is widely studied. \citet{zhang2024investigating} propose a framework to study the contribution of the individual layers by performing an ablation study on layer exclusion via an adapted framework of Shapley values. Results suggest that early layers, called ``cornerstone'' layers, process initial embedding into the space where subsequent ``non-cornerstone'' layers operate. Such ``non-cornerstone'' layers can overlap in function, and their individual contribution is not significant. \citet{gromov2025the} makes an argument that deeper layers do not store knowledge, but rather are utilized for the computations, e.g., reasoning and dealing with long context. Similar to prior work, they studied model performance by ablating model layers and optionally fine-tuning. Further works show that it is possible to improve computational complexity and/or performance by modifying the flow of the embedding in the model~\citep{heakl2025drllmdynamiclayerrouting,laitenberger2025layerswhenlearningskip,li2025skiplayerloopit}.
\textbf{Layers as Painters} is a particular framework by \citet{sun2025layesaspainters} that inspired our work. It was originally designed to verify the hypothesis that different layers express shared representational ``languages'' by operating in a common representation space and complement each other's output. This contrasts with operating in a hierarchical feature space, similar to models with bottlenecks such as convolutional networks. If this holds, then the sequence of the trained layers can be reordered as illustrated in Figure~\ref{fig:layer_experiments} without catastrophic loss of performance.

Lastly, we complement this framework with \textbf{probing classifiers} \citep{belinkov2022probing}, where a classification model is trained on an embedding produced by a model (or intermediate stage) that is studied to infer some (e.g., linguistic) quantity, as illustrated in Figure~\ref{fig:probing_cls}. The performance of this probing classifier gives a rough estimation of the mutual information between the embedding and the quantity of interest. For \TabPFNtwo{}, a fine-tuned decoder can be used to early-exit the forward pass and reduce inference time while maintaining performance~\citep{Kuken2025EarlyStopping}, suggesting that not all layers are equally important; however, it remains open to what extent this transfers across models. Additionally, \citet{ye2025closer} analyzed the embeddings of each layer for \TabPFNtwo{} using PCA.

\section{Methodology and Experiment Design}
\label{sec:method}

\begin{figure}[tbp]
    \centering
    \vspace{-0.5cm}
    \captionsetup[subfigure]{aboveskip=0pt, belowskip=0pt} 
    \begin{subfigure}[b]{0.24\textwidth}
        \centering
        \includegraphics[height=2cm]{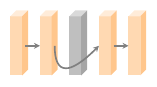}
        \caption{Skipping layers}
        \label{fig:skip_layer}
    \end{subfigure}
    \hfill
    \begin{subfigure}[b]{0.24\textwidth}
        \centering
        \includegraphics[height=2cm]{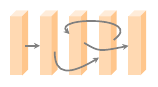}
        \caption{Swapping layers}
        \label{fig:swap_layer}
    \end{subfigure}
    \hfill
    \begin{subfigure}[b]{0.24\textwidth}
        \centering
        \includegraphics[height=2cm]{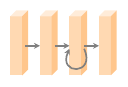}
        \caption{Repeating layers}
        \label{fig:repeat_layer}
    \end{subfigure}
    \begin{subfigure}[b]{0.24\textwidth}
        \centering
        \includegraphics[height=2cm]{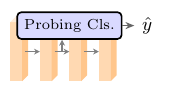}
        \caption{{Probing Classifier}}
        \label{fig:probing_cls}
    \end{subfigure}
    \caption{Layer reorganization in tabular ICL models following the ``Layers as Painters'' framework.}
    \label{fig:layer_experiments}
\end{figure}
In this section, we formulate the objectives of our structured analysis and describe each of the experiments. Namely, we focus on three guiding questions:

\textbf{Q1: Do all layers speak the same language?}  
We analyze whether the application of the layers is commutative or whether there is hierarchy in the representations. 
First, to test representational alignment, we swap the order of layers and measure the performance of the forward pass, as shown in Figure~\ref{fig:swap_layer}. If switching layers degrades performance, the representation might have a hierarchy. Secondly, we perform probing with linear classifiers trained on the representation from the same and other layers. Representations are similar if a trained probing classifier performs well on another layer's embedding. Thirdly, we study repeating layers as shown in Figure~\ref{fig:repeat_layer}. If repeating a layer increases performance, it might suggest that the layer performs
iterative refinement, e.g., as in recurrent architectures. This also means that the layers' in- and output embedding spaces must be closely aligned for the layer to operate in a recurrent manner \citep{dehghani2018universal}. 

\textbf{Q2: Does the model use all layers?}  
To study this, we skip layers in the forward pass (see Figure~\ref{fig:skip_layer}) and measure the impact on downstream performance. A negligible change in performance might indicate that a layer is redundant (or its contribution is not required to solve the task).

\textbf{Q3: How consistent are findings across models?} Finally, to assess whether our findings are specific to a single instantiation of a tabular ICL model or apply to multiple models, we repeat the analysis for \textsc{TabPFN-v1}, \textsc{TabPFN-v2}, and \textsc{TabICL}.

\textbf{Models and Datasets:} We run experiments on a subset of TabArena comprising 15 binary classification tasks \citep{erickson2025tabarena} (see Appendix~\ref{app:datasets}). We use \TabPFNone{}, \TabPFNtwo{}, and \TabICL{} as tabular ICL models. We use ROC-AUC, averaged across datasets, as the evaluation metric.

\textbf{Classifier probing.}
For this task, we use three different models: logistic regression, K-Nearest Neighbors (KNN), and a fine-tuned MLP from the respective ICL model's decoder layer. We begin by extracting embeddings for each layer from its hidden states in the ICL setup. Specifically, we train and evaluate the probing classifier only on the embedded query set. For this experiment, we extend the query set by including half of the original training set (excluded from the support set) to serve as training data for the probing classifier.

\section{Experimental Results}

\begin{figure}[t]
\centering
\vspace{-0.2cm}
\includegraphics[width=0.8\textwidth]{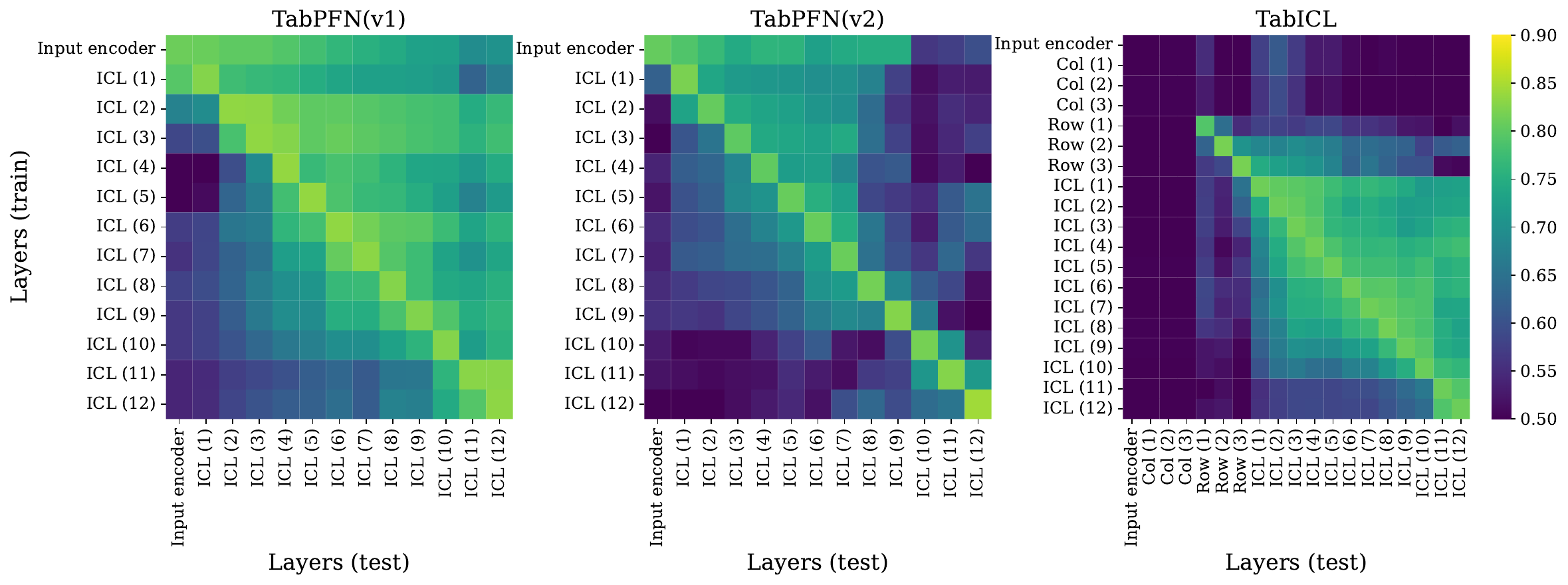}
\vspace{-0.2cm}
\caption{Average AUC for probing classifiers (logistic regression) trained on embeddings at different layers of the models.} 
\label{fig:linear_probing}
\end{figure}

\begin{figure}[htbp]
\centering
\includegraphics[height=3.5cm]{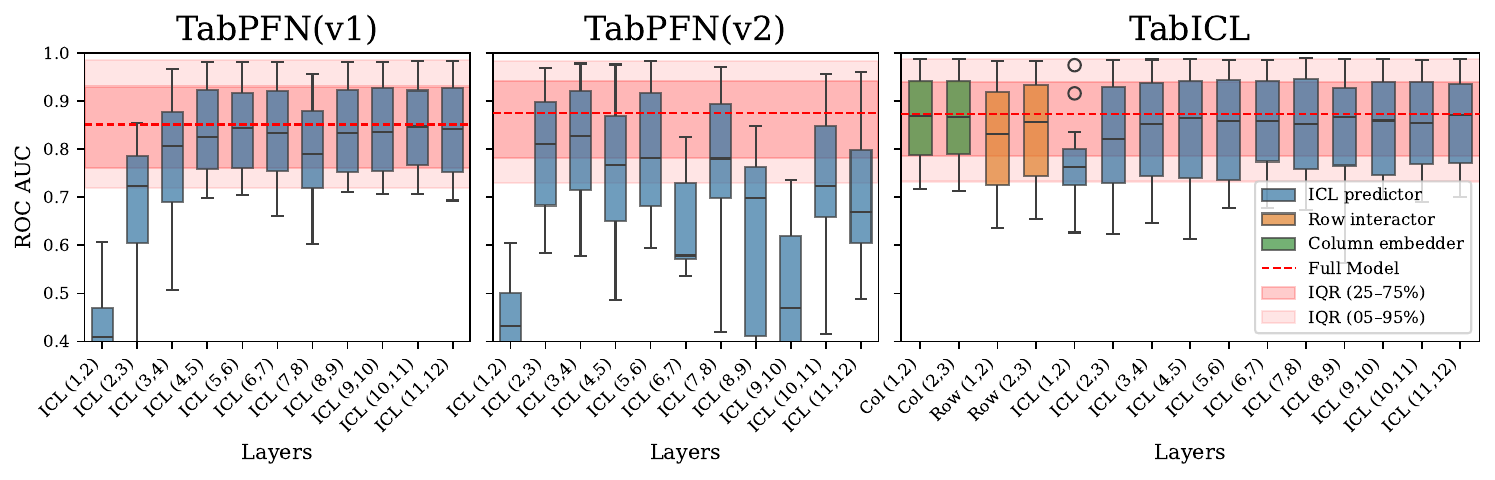}
\vspace{-0.2cm}
\caption{Impact on average AUC when swapping two layers in the forward pass of the models. } 
\label{fig:swap_layers}
\end{figure}

\begin{figure}[htbp]
\centering
\includegraphics[height=3.5cm]{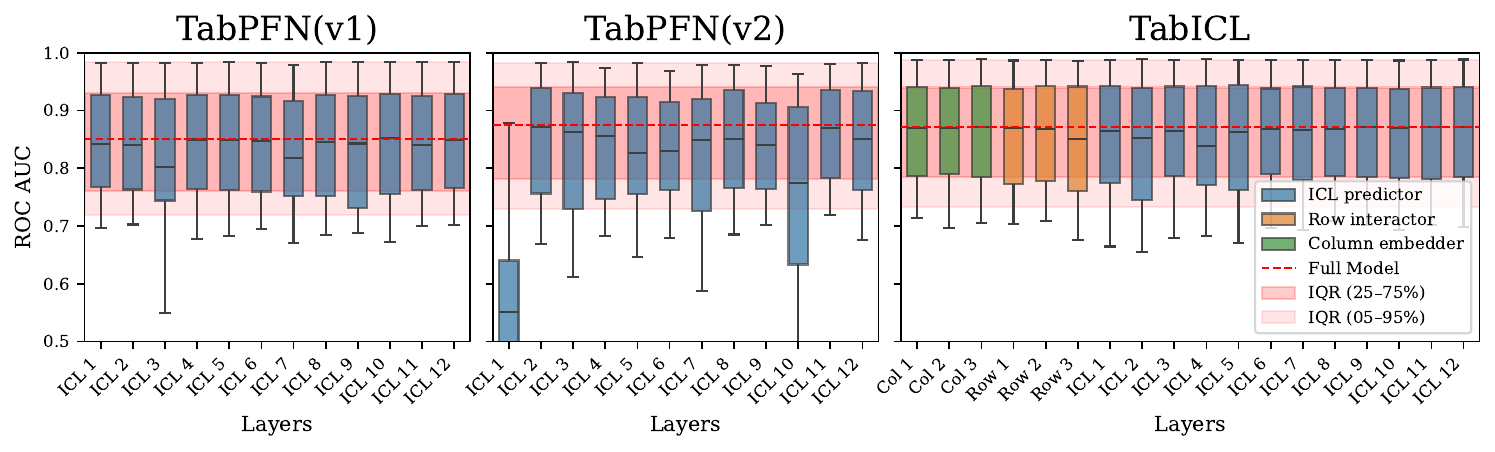}
\vspace{-0.2cm}
\caption{Impact of repeating the layers of the models. } 
\label{fig:repeating_layers}
\end{figure}

\textbf{Q1: Some layers do speak the same language.} The \textbf{probing} results in Figure~\ref{fig:linear_probing} indicate that while the outcomes are highly model-dependent, a consistent pattern emerges:  A probing classifier trained on layer $i$ performs well on the embeddings of a later layer $j>i$, whereas the reverse is not true. This suggests that later layers still contain information from earlier layers, and that new features emerge in higher layers not present in earlier ones. This is particularly pronounced for \TabPFNtwo{}. For \TabPFNone{}, this is only noticeable for early layers, whereas for \TabICL{}, this behaviour is least pronounced. These results support the ``layers as painters'' hypothesis that the middle layers share a similar representational space, as probes trained at different layers can operate interchangeably. In Appendix~\ref{app:probing}, we also provide concise similarity analyses between the embeddings of each layer and include the results of using a KNN classifier and a fine-tuned model decoder as probes. 
Figure~\ref{fig:swap_layers} shows the effect of \textbf{layer swapping}. As expected, performance degrades most in the first (or early layers), indicating that layer order is important.
Interestingly, for \TabPFNtwo{}, there is a significant loss in the middle layers, suggesting that while these layers may share a similar representation space, they perform different operations. 
Finally, we study \textbf{repeating individual layers} in Figure \ref{fig:repeating_layers}. Interestingly, and in contrast to LLMs, where repeating a single layer is highly detrimental~\citep{sun2025layesaspainters}, in tabular models, repeating a layer seems less harmful or can even improve performance for some tasks (see Figure~\ref{app:fig:wtl_repeating_layers}).

\begin{figure}[htbp]
\centering
\includegraphics[height=3.5cm]{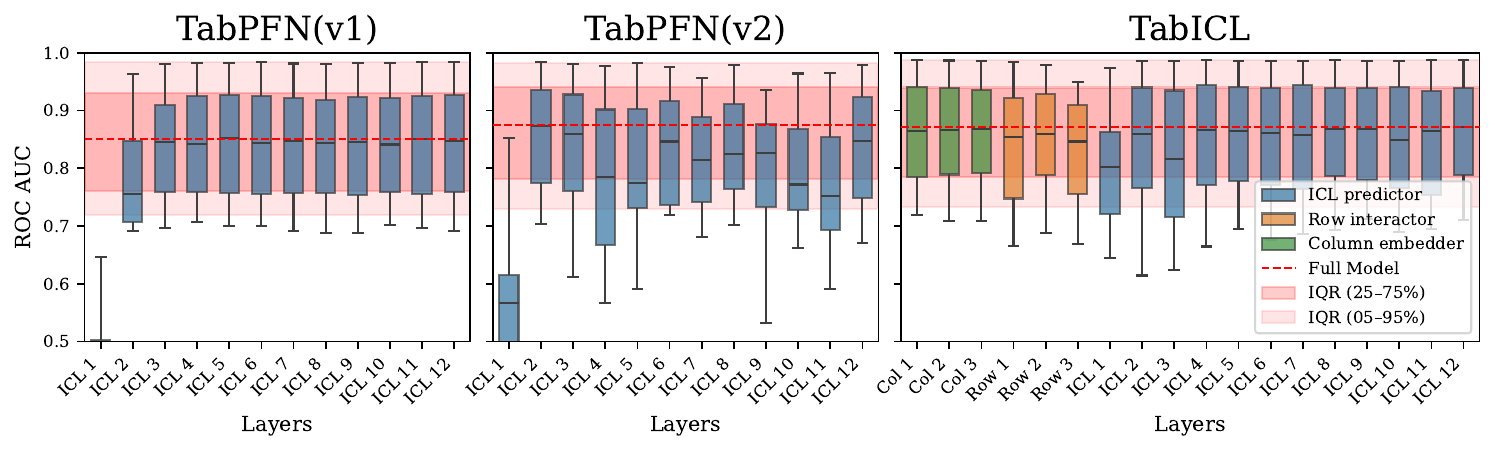}
\vspace{-0.2cm}
\caption{Impact on average AUC when skipping a layer during the forward pass.} 
\label{fig:skipping_layers}
\end{figure}

\textbf{Q2: Yes, some layers can be skipped.} 
Figure~\ref{fig:skipping_layers} shows that skipping early layers of the ICL prediction stage of \TabICL{} and \TabPFNone{} impacts performance most. This means that earlier layers are more important for final performance than later ones (and final layers can even be dropped with minimal loss). 
This is in line with findings from the LLM literature \citep{zhang2024investigating,sun2025layesaspainters}. Results are different for \TabPFNtwo{}; multiple intermediate layers appear essential for final performance. Prompted by these findings, we test early exiting like \cite{Kuken2025EarlyStopping} in Appendix~\ref{app:eraly_exit}. Our results indicate that for \TabPFNone{} and \TabICL{}, early exit is feasible without fine-tuning the decoder components. In contrast, \TabPFNtwo{} requires an individually fine-tuned decoder, as also observed by \citet{Kuken2025EarlyStopping}.

\textbf{Q3: Some common patterns do emerge; the results vary greatly.} Our results suggest that early layers have the most impact for all models. Unlike in LLMs, \TabPFNtwo{} seems to have less redundancy according to our probing and layer skipping results. Also, repeating a layer often neither degrades nor improves performance, suggesting that the layer has learned to modulate its contribution to the output as needed. Overall results on specific layers are model-specific. Additionally, in Appendix~\ref{app:layer_reorganization}, we also provide win–tie–lose results for our experiments, which suggest that performance changes caused by each layer reorganization operation are also task-dependent.

\section{Conclusion and Discussion}
\label{sec:discussion}
Our analysis highlights fundamental differences between tabular ICL models and LLM layer dynamics. We find that not all layers in tabular ICL models contribute equally, with some redundancy across depth. Specifically, we observed that later layers in \TabICL{} and \TabPFNone{} are not critical to the final performance. 
We note that our evaluation averages effects over datasets using a single fold, repetition, and model initialization, and our layer-reorganization experiments modify only one layer at a time. However, even within these limitations, our empirical study 
raises promising questions for future research: How stable are these observations across (1) tasks and (2) initializations \citep{wang2018towards}? And more broadly, can these insights guide the development of (3) lightweight and (4) interpretable tabular ICL models? 

\begin{ack}
This research has been funded by the Federal Ministry of Research, Technology and Space of Germany and the state of North Rhine-Westphalia as part of the Lamarr Institute for Machine Learning and Artificial Intelligence. Additionally, this research utilized compute resources at the Tübingen Machine Learning Cloud, DFG FKZ INST 37/1057-1 FUGG.  A. Balef and M. Koshil also thank the International Max Planck Research School for Intelligent Systems (IMPRS-IS).
\end{ack}

%--------------------------------------------->Bibliography
\bibliographystyle{unsrtnat}
\bibliography{strings,references,lib,proc, myproc}

@inproceedings{hollmann-iclr23a,
  title        = {Tab{PFN}: A Transformer That Solves Small Tabular Classification Problems in a Second},
  author       = {N. Hollmann and S. M{\"u}ller and K. Eggensperger and F. Hutter},
  crossref     = {iclr23},
}

@article{hollmann-nature25a,
  title={Accurate predictions on small data with a tabular foundation model},
  author={N. Hollmann and S. M{\"u}ller and L. Purucker and A. Krishnakumar and M. K{\"o}rfer and Shi Bin Hoo and Robin Tibor Schirrmeister and Frank Hutter},
  journal={Nature},
  volume={637},
  number={8045},
  pages={319--326},
  year={2025},
  publisher={Nature Publishing Group UK London}
}

@proceedings{aaai25,
  title        = {Proceedings of the Thirty-Eighth Conference on Artificial Intelligence ({AAAI}'25)},
  year         = 2025,
  booktitle    = {Proceedings of the Thirty-Eighth Conference on Artificial Intelligence ({AAAI}'25)},
  publisher    = {{AAAI} Press},
  organization = {Association for the Advancement of Artificial Intelligence},
}

@proceedings{neuripsdbt25,
  title        = {Proceedings of the Neural Information Processing Systems Track on Datasets and Benchmarks},
  year         = 2025,
  booktitle    = {Proceedings of the Neural Information Processing Systems Track on Datasets and Benchmarks},
  publisher = curran,
}

@proceedings{neurips25,
  title = {Proceedings of the 38th International Conference on Advances in Neural Information Processing Systems ({N}eur{IPS}'25)},
  booktitle = {Proceedings of the 38th International Conference on Advances in Neural Information Processing Systems ({N}eur{IPS}'25)},
  publisher = curran,
  year = {2025},
}

@proceedings{neurips18m,
  title        = {Proceedings of the 31st International Conference on Advances in Neural Information Processing Systems ({N}eur{IPS}'18)},
  year         = 2018,
  booktitle    = {Proceedings of the 31st International Conference on Advances in Neural Information Processing Systems ({N}eur{IPS}'18)},
  publisher    = curran,
}

@proceedings{neurips20m,
  title        = {Proceedings of the 33rd International Conference on Advances in Neural Information Processing Systems ({N}eur{IPS}'20)},
  year         = 2020,
  booktitle    = {Proceedings of the 33rd International Conference on Advances in Neural Information Processing Systems ({N}eur{IPS}'20)},
  publisher    = curran,
}

@proceedings{icml24m,
  title = {Proceedings of the 41st International Conference on Machine Learning ({ICML}'24)},
  booktitle = {Proceedings of the 41st International Conference on Machine Learning ({ICML}'24)},
  publisher    = {PMLR},
  series       = {Proceedings of Machine Learning Research},
  year = {2024},
  volume       = 251,
}

@proceedings{iclr19,
  title        = {Proceedings of the International Conference on Learning Representations ({ICLR}'19)},
  year         = 2019,
  booktitle    = {The Seventh International Conference on Learning Representations ({ICLR}'19)},
  organization = {ICLR},
}

@proceedings{iclr23,
  title        = {Proceedings of the International Conference on Learning Representations ({ICLR}'23)},
  year         = 2023,
  booktitle    = {The Eleventh International Conference on Learning Representations ({ICLR}'23)},
  organization = {ICLR},
}

@proceedings{iclr25,
  title        = {Proceedings of the International Conference on Learning Representations ({ICLR}'25)},
  year         = 2025,
  booktitle    = {The Thirteenth International Conference on Learning Representations ({ICLR}'25)},
  organization = {ICLR},
}

@inproceedings{brown-neurips20,
  title        = {Language Models are Few-Shot Learners},
  author       = {T. Brown and B. Mann and N. Ryder and M. Subbiah and J. Kaplan and P. Dhariwal and A. Neelakantan and P. Shyam and G. Sastry and A. Askell and S. Agarwal and A. Herbert-Voss and G. Krueger and T. Henighan and R. Child and A. Ramesh and D. Ziegler and J. Wu and C. Winter and C. Hesse and M. Chen and E. Sigler and M. Litwin and S. Gray and B. Chess and J. Clark and C. Berner and S. McCandlish and A. Radford and I. Sutskever and D. Amodei},
  pages        = {1877--1901},
  crossref     = {neurips20m},
}

@inproceedings{qu2025tabicl,
  title        = {Tab{ICL}: A Tabular Foundation Model for In-Context Learning on Large Data},
  author       = {J. Qu and D. Holzmüller and G. Varoquaux and M. Le Morvan},
  crossref     = {icml24m},
}

@inproceedings{sun2025layesaspainters,
  author       = {Q. Sun and M. Pickett and A. K. Nain and L. Jones},
  title        = {Transformer layers as painters},
  crossref     = {aaai25},
}

@inproceedings{gromov2025the,
  title        = {{The Unreasonable Ineffectiveness of the Deeper Layers}},
  author       = {A. Gromov and K. Tirumala and H. Shapourian and P. Glorioso and D. Roberts},
  crossref     = {iclr25},
}

@inproceedings{erickson2025tabarena,
  title        = {{TabArena}: A Living Benchmark for Machine Learning on Tabular Data},
  author       = {N. Erickson and L. Purucker and A. Tschalzev and D. Holzmüller and P. M. Desai and D. Salinas and F. Hutter},
  crossref     = {neuripsdbt25},
}

@inproceedings{ye2025closer,
  title        = {A closer look at {TabPFN} v2: Understanding its strengths and extending its capabilities},
  author       = {H. J. Ye and S. Y. Liu and W. L. Chao},
  crossref     = {neurips25},
}

@inproceedings{wang2018towards,
  title        = {Towards understanding learning representations: To what extent do different neural networks learn the same representation},
  author       = {L. Wang and L. Hu and J. Gu and Z. Hu and Y. Wu and K. He and J. Hopcroft},
  crossref     = {neurips18m},
}

@inproceedings{dehghani2018universal,
  title        = {Universal Transformers},
  author       = {M. Dehghani and Ç. Gülçehre and Y. Bengio and others},
  crossref     = {iclr19},
}

@article{laitenberger2025layerswhenlearningskip,
  title        = {What Layers When: Learning to Skip Compute in {LLMs} with Residual Gates},
  author       = {F. Laitenberger and D. Kopiczko and C. G. M. Snoek and Y. M. Asano},
  year         = {2025},
  eprint       = {2510.13876},
  archivePrefix = {arXiv},
  primaryClass  = {cs.CL},
  journal      = {arXiv preprint},
  url          = {https://arxiv.org/abs/2510.13876},
}

@article{li2025skiplayerloopit,
  title        = {Skip a Layer or Loop it? Test-Time Depth Adaptation of Pretrained {LLMs}},
  author       = {Z. Li and Y. Li and T. Zhou},
  year         = {2025},
  eprint       = {2507.07996},
  archivePrefix = {arXiv},
  primaryClass  = {cs.LG},
  journal      = {arXiv preprint},
  url          = {https://arxiv.org/abs/2507.07996},
}

@article{heakl2025drllmdynamiclayerrouting,
  title        = {{Dr.LLM}: Dynamic Layer Routing in {LLMs}},
  author       = {A. Heakl and M. Gubri and S. Khan and S. Yun and S. J. Oh},
  year         = {2025},
  eprint       = {2510.12773},
  archivePrefix = {arXiv},
  primaryClass  = {cs.CL},
  journal      = {arXiv preprint},
  url          = {https://arxiv.org/abs/2510.12773},
}

@inproceedings{zhang2024investigating,
  title        = {Investigating Layer Importance in Large Language Models},
  author       = {Y. Zhang and Y. Dong and K. Kawaguchi},
  booktitle    = {The 7th BlackboxNLP Workshop},
  year         = {2024},
  url          = {https://openreview.net/forum?id=kjZIIvFtmK},
}

@article{belinkov2022probing,
  author       = {Y. Belinkov},
  title        = {Probing Classifiers: Promises, Shortcomings, and Advances},
  journal      = {Computational Linguistics},
  year         = {2022},
  doi          = {10.1162/coli_a_00422},
  url          = {https://doi.org/10.1162/coli_a_00422},
}

@inproceedings{Kuken2025EarlyStopping,
  author       = {J. Küken and L. Purucker and F. Hutter},
  title        = {Early Stopping Tabular In-Context Learning},
  booktitle    = {1st International Workshop on Foundation Models for Structured Data (FMSD) @ ICML 2025},
  year         = {2025},
}

@STRING{aaai    = "Proceedings of the National Conference on Artificial
                  Intelligence (AAAI)" }

@STRING{curran  = "Curran Associates"}

@STRING{nature = "Nature"}

@STRING{pmlr    = "Proceedings of Machine Learning Research"}
%%%%%%%%%%%%%%%%%%%%%%%%%%%%%%%%%%%%%%%%%%%%%%%%%%%%%%%%%%%%
\newpage
\appendix

\section{Datasets}
\label{app:datasets}
Table~\ref{app:tab:dataset_list} lists all datasets used in our experiments.
\begin{table}[htbp]
\caption{Dataset.}
\label{app:tab:dataset_list}
\begin{tabular}{lrlrrr}
\toprule
index & task id & dataset name & \#samples & \#features & \#categorical features \\
\midrule
1 & 363619 & Bank-Customer-Churn & 10000 & 11 & 5 \\
2 & 363621 & blood-transfusion-service-center & 748 & 5 & 1 \\
3 & 363623 & churn & 5000 & 20 & 5 \\
4 & 363624 & coil2000-insurance-policies & 9822 & 86 & 4 \\
5 & 363626 & credit-g & 1000 & 21 & 14 \\
6 & 363629 & diabetes & 768 & 9 & 1 \\
7 & 363671 & Fitness-Club & 1500 & 7 & 4 \\
8 & 363674 & hazelnut-spread-contaminant-detection & 2400 & 31 & 1 \\
9 & 363682 & Is-this-a-good-customer & 1723 & 14 & 9 \\
10 & 363684 & Marketing-Campaign & 2240 & 26 & 9 \\
11 & 363689 & NATICUSdroid & 7491 & 87 & 87 \\
12 & 363694 & polish-companies-bankruptcy & 5910 & 65 & 1 \\
13 & 363696 & qsar-biodeg & 1054 & 42 & 6 \\
14 & 363700 & seismic-bumps & 2584 & 16 & 4 \\
15 & 363706 & taiwanese-bankruptcy-prediction & 6819 & 95 & 1 \\
\bottomrule
\end{tabular}
\end{table}

\clearpage
\section{More results}
\label{app:more_results}

\subsection{Probing}
\label{app:probing}
\begin{figure}[tbp]
\centering
\includegraphics[width=1.0\textwidth]{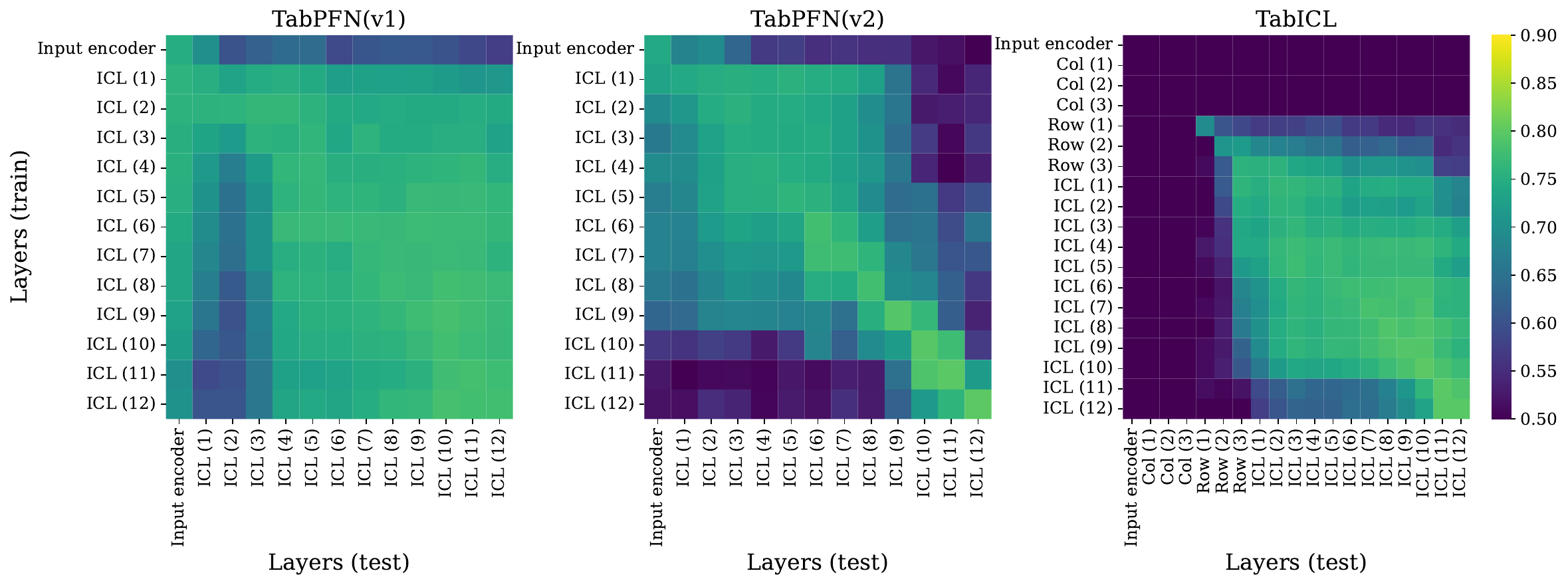}
\caption{Probing with KNN trained on embedding from different layers of the models. } 
\label{app:fig:knn_linear_probing}
\end{figure}

\begin{figure}[tbp]
\centering
\includegraphics[width=1.0\textwidth]{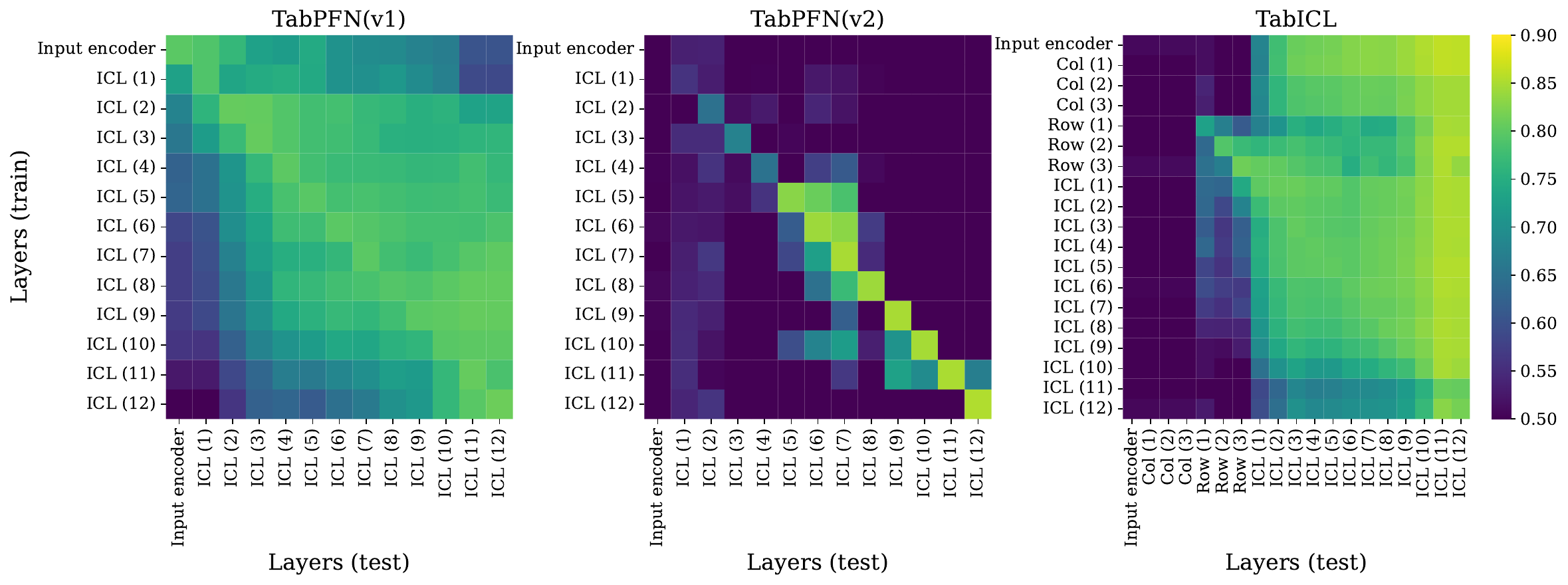}
\caption{Probing with model decoder fine-tuned on embedding from different layers of the models. } 
\label{app:fig:model_decoder_linear_probing}
\end{figure}

We provide more results from linear probing using different classifiers. Figure~\ref{app:fig:knn_linear_probing} shows the performance of KNN trained on embeddings extracted from different layers of the models, while Figure~\ref{app:fig:model_decoder_linear_probing} presents results from model decoders fine-tuned on the same embeddings.

\begin{figure}[tbp]
\centering
\includegraphics[width=1.0\textwidth]{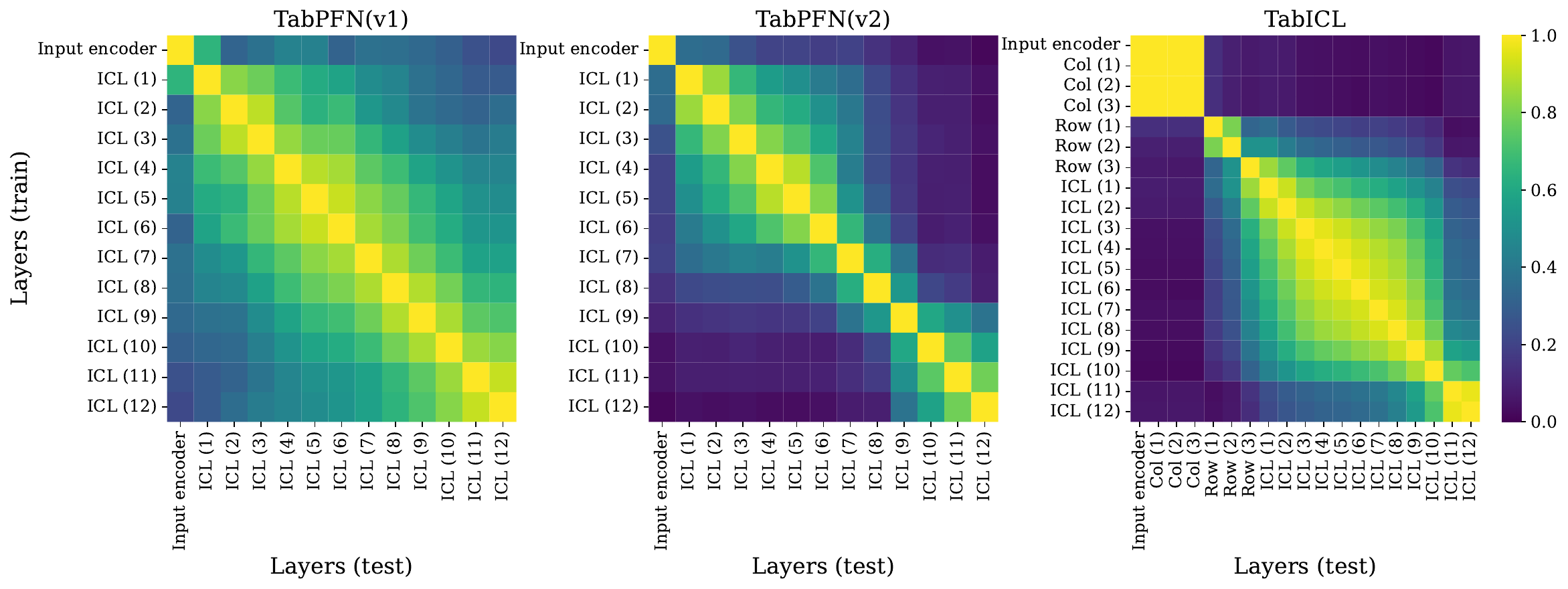}
\caption{Cosine similarity of embeddings extracted from different layers of the models.}
\label{app:fig:cosine_similarity}
\end{figure}

We also report the cosine similarity between embeddings from different layers of the models, as shown in Figure~\ref{app:fig:cosine_similarity}.

\subsection{Layer reorganization}
\label{app:layer_reorganization}

\begin{figure}[tbp]
\centering
\includegraphics[width=\textwidth]{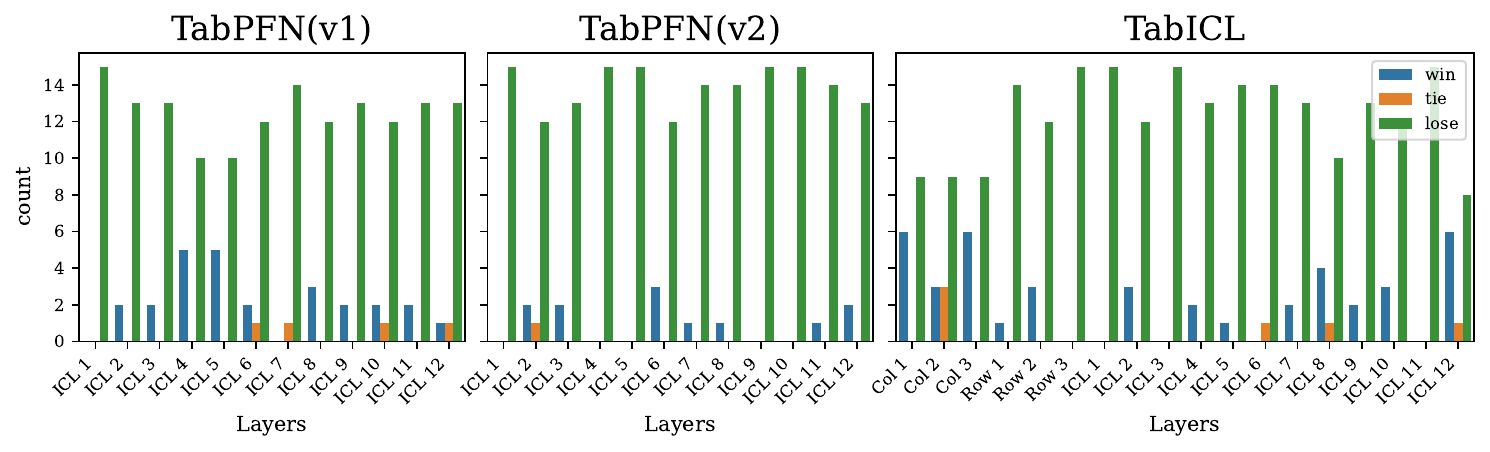}
\caption{Effect of skipping layers in the tabular foundation model's architecture.}
\label{app:fig:wtl_skipping_layers}
\end{figure}

\begin{figure}[tbp]
\centering
\includegraphics[width=\textwidth]{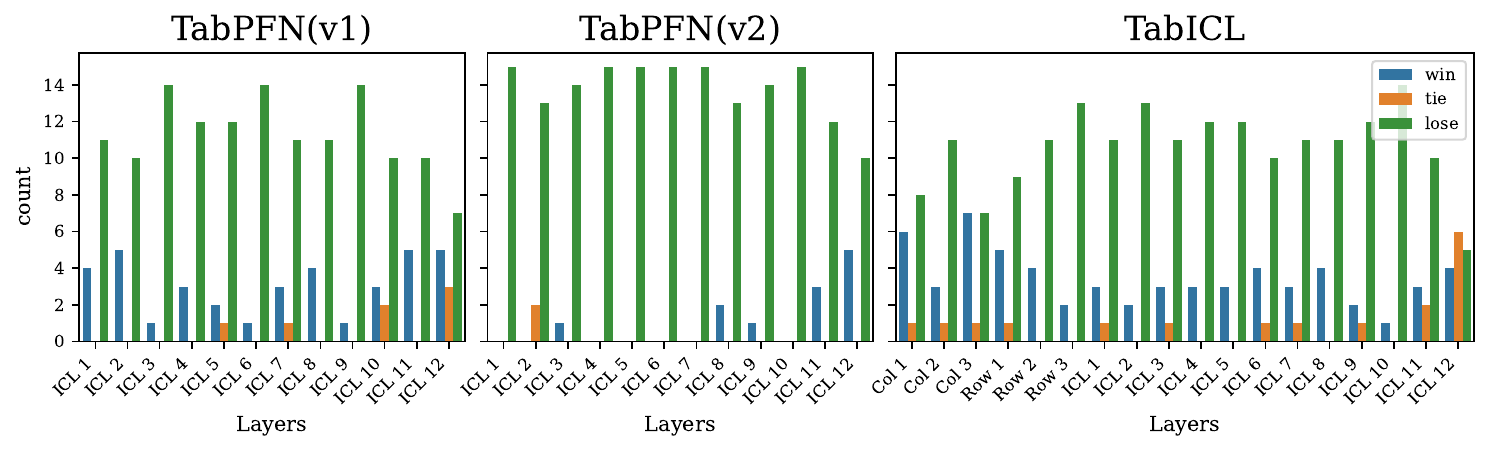}
\caption{Effect of repeating layers in the tabular foundation model's architecture.}
\label{app:fig:wtl_repeating_layers}
\end{figure}

\begin{figure}[tbp]
\centering
\includegraphics[width=\textwidth]{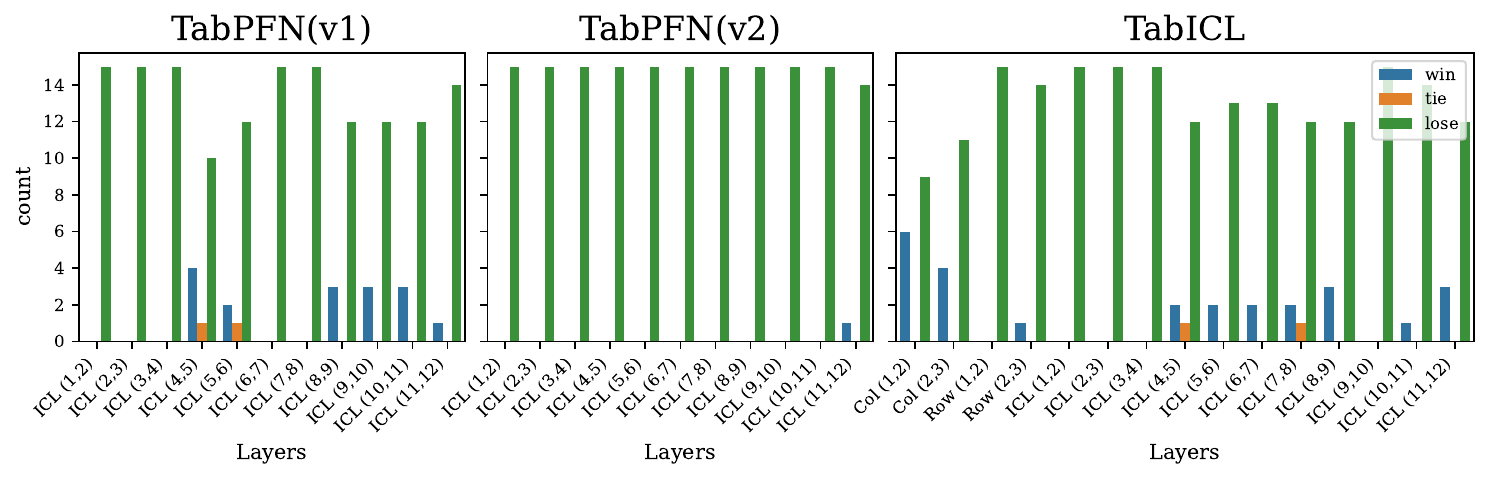}
\caption{Effect of swapping layers in the tabular foundation model's architecture.}
\label{app:fig:wtl_swap_layers}
\end{figure}

Here, we provide win–tie–lose comparison plots against the full model evaluation to examine how architectural changes in the transformer affect performance across tasks. We adopt a tie threshold of $2 \times 10^{-4}$ to ensure that numerical precision and minor fluctuations do not result in spurious wins or losses. Figure~\ref{app:fig:wtl_skipping_layers} shows the effect of skipping layers, Figure~\ref{app:fig:wtl_repeating_layers} illustrates the impact of repeating layers, and Figure~\ref{app:fig:wtl_swap_layers} presents the results of swapping layers.

\subsection{Early exit}
\label{app:eraly_exit}

\begin{figure}[tbp]
\centering
\includegraphics[width=\textwidth]{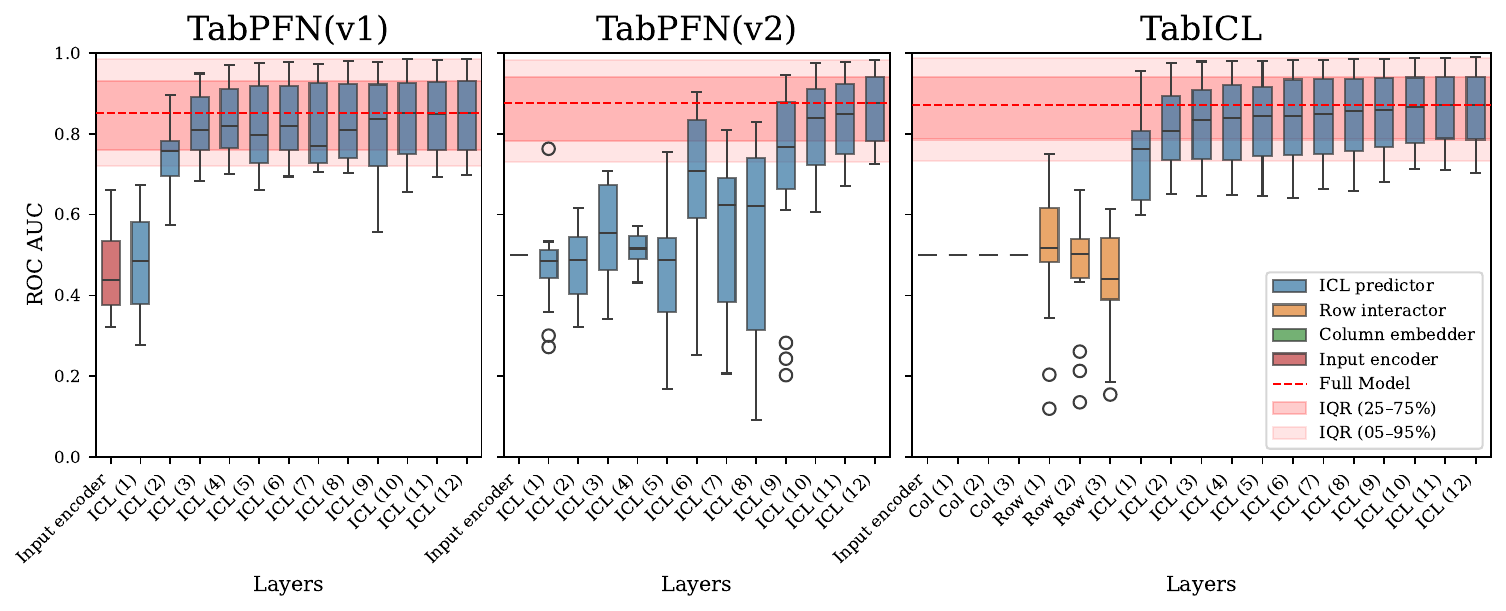}
\caption{Effect of the early exit strategy on the tabular foundation model's performance.}
\label{app:fig:early_exit}
\end{figure}

\begin{figure}[tbp]
\centering
\includegraphics[width=\textwidth]{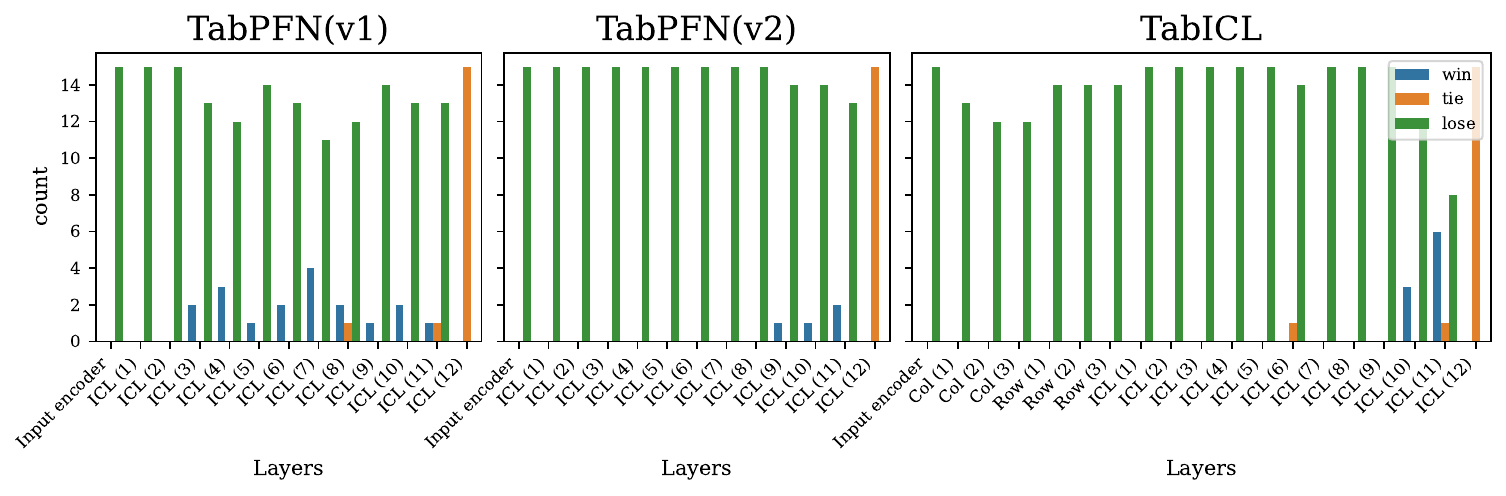}
\caption{Comparison of performance with and without the early exit strategy in the tabular foundation model.}
\label{app:fig:wtl_early_exit}
\end{figure}

We perform an early-exit strategy, meaning that after each layer, we pass the embeddings to the decoder. However, we did not fine-tune the decoder layer in contrast to \cite{Kuken2025EarlyStopping}. As observed in Figure \ref{app:fig:early_exit} and Figure \ref{app:fig:wtl_early_exit} for \TabPFNtwo{}, having an individual decoder is necessary, as shown by \citet{Kuken2025EarlyStopping}, whereas for other models, the performance degradation is not drastic.

\end{document}